\documentclass{article}
\usepackage{spconf,amsmath,graphicx}

\usepackage{color}
\usepackage{enumitem}
\usepackage{multirow}
\usepackage{etoolbox,siunitx}
\robustify\bfseries
\usepackage{booktabs}
\usepackage{xcolor}
\usepackage{tcolorbox}
\usepackage{url}
\usepackage{amsmath}
\usepackage{graphicx}
\usepackage{balance}
\usepackage{amssymb}
\usepackage{bm}
\usepackage{arydshln}
\usepackage{caption}
\usepackage{cite}
\usepackage{hyperref}
\usepackage{colortbl}

\title{Evolutionary Prompt Design for LLM-Based Post-ASR Error Correction}
\name{Rithik Sachdev$^1$\quad Zhong-Qiu Wang$^2$\quad Chao-Han Huck Yang$^3$}
\address{$^1$Carnegie Mellon University \quad $^2$Southern University of Science and Technology \quad $^3$NVIDIA Research }
\begin{document}
\maketitle
\begin{abstract}
Building upon the strength of modern large language models (LLMs), generative error correction (GEC) has emerged as a promising paradigm that can elevate the performance of modern automatic speech recognition (ASR) systems.
One representative approach is to leverage in-context learning to prompt LLMs so that a better hypothesis can be generated by the LLMs based on a carefully-designed prompt and an $N$-best list of hypotheses produced by ASR systems.
However, it is yet unknown whether the existing prompts are the most effective ones for the task of post-ASR error correction.
In this context, this paper first explores alternative prompts to identify an initial set of effective prompts, and then proposes to employ an evolutionary prompt optimization algorithm to refine the initial prompts.
Evaluations results on the CHiME-4 subset of the Task $1$ of the SLT $2024$ GenSEC challenge show the effectiveness and potential of the proposed algorithms.\footnote{Our code is open sourced at \url{https://github.com/rithiksachdev/PostASR-Correction-SLT2024}. }
\end{abstract}
\begin{keywords}
Evolutionary prompt optimization, post-ASR error correction, large language models, automatic speech recognition.
\end{keywords}
\section{Introduction}\label{intro}

Large language models (LLMs) \cite{Minaee2024}, trained on massive textual data using neural network-based transformer architectures, have revolutionized many tasks in natural language processing. The auto-regressive learning nature of LLMs introduces and empowers a new ``\textit{prompting}'' mechanism based on input instructions, where users can provide text prompts to guide LLMs to complete particular tasks \cite{Liu2022} as a form of next token prediction.

One of these popular prompt-activated tasks is post-automatic speech recognition (ASR) text correction, where based on an $N$-best list of hypotheses generated by ASR systems, the task is to predict the true transcription \cite{Chen2023HyPoradise, Radhakrishnan2023, Gu2024DenoisingLM, Hu2024robustASR}.
Although this problem can be approached by language model re-scoring techniques~\cite{yang2021multi,liu2016attention, ma2023n}, recent studies \cite{Chen2023HyPoradise, Radhakrishnan2023, Gu2024DenoisingLM, Hu2024robustASR} have shown that leveraging LLMs to correct errors in an generative way often leads to better performance. Specifically, LLM-based generative error correction~\cite{yang2023generative} (GER) can infer phonetic similarity and contextual information to reduce errors beyond oracle ranking result of the $N$-hypotheses list.

When \textit{input prompts} are critical for instructing LLMs on unseen ASR tasks, these task-activating prompts \cite{yang2023generative} are often empirically-designed and under-explored in the research community. For example, early works~\cite{watanabe2017language, yang2021voice2series,gao2022wavprompt, chang2023speechprompt} focus on iterative optimization at the waveform level to instruct acoustic models for new tasks, but there are fewer studies on optimizing LLM-prompts for ASR tasks.

To achieve better post-ASR error correction, this paper explores alternative prompts for this task, and investigates a conditional evolutionary strategies based prompt optimization algorithm, named EvoPrompt \cite{Guo2024Evoprompt}, to refine the alternative prompts.
Evaluation results on the CHiME-4 subset of the HyPoradise dataset \cite{Chen2023HyPoradise} show the effectiveness of the proposed algorithms.

In the rest of this paper, we describe the proposed algorithm in Section \ref{proposed_algorithm}, experimental setup in Section \ref{exp_setup}, evaluation results in Section \ref{results}, and conclusions in Section \ref{conclusions}.

\section{Proposed Algorithms}\label{proposed_algorithm}

\begin{figure}
  \centering  
  \includegraphics[width=8.5cm]{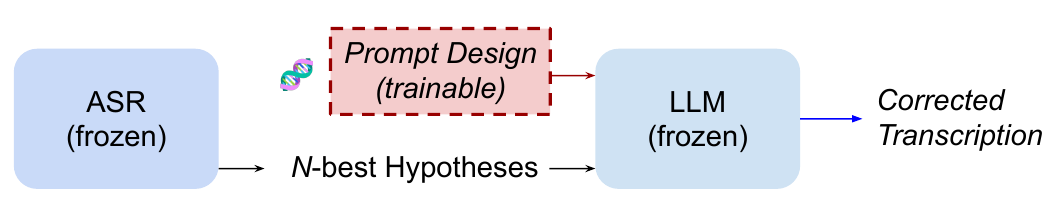}
  \caption{Approach overview, where an $N$-best list of hypotheses and a trainable prompt instruction are fed to a pre-trained LLM for error correction. Details of prompt design processes are shown in Fig. \ref{in_context_learning_example}.}
  \label{system_figure}
\end{figure}

Fig. \ref{system_figure} illustrates the proposed system, where an $N$-best list of hypotheses and a prompt are input to an LLM for error correction.
This section first describes several empirically-designed alternative prompts, and then leverages an evolutionary algorithm named EvoPrompt \cite{Guo2024Evoprompt} for prompt optimization.

\subsection{Alternative Prompt Design}\label{alterative_prompts}

In Task $1$ of the GenSEC challenge, the default prompt provided by the challenge organizers is \textit{You need to do language model rescoring in ASR. Given the 5-best hypotheses, you need to report the true transcription from the 5-best hypotheses.} We denote this prompt as \textbf{Baseline Prompt}.
It has shown strong performance in post-ASR error correction tasks \cite{Chen2023HyPoradise}, but it is unknown whether there exist better prompts.
To improve the performance, we propose the following alternative prompts:
\begin{itemize}[leftmargin=*,noitemsep,topsep=0pt]
\item \textbf{Prompt \#1}: \textit{This is a hard problem. You need to report the true transcription from the 5-best hypotheses.}.
\item \textbf{Prompt \#2}: \textit{This is a hard problem. Carefully summarize in ONE detailed sentence the following captions by different people describing the same audio. Do not allude to the existence of the multiple captions. Focus on describing the content of the audio. The transcriptions have some intialism for corporations.}
\item \textbf{Prompt \#3}: \textit{You have been given 5 different possible captions of the same audio, carefully summarize the given text in one sentence. Make sure to follow all the steps in identifying the right summary.}
\item \textbf{Prompt \#4}: \textit{There is some financial data discussed in a meeting. You need to correctly give the true transcription from the 5-best hypotheses. Your task is to critically evaluate these option using english grammar. Mostly these transcriptions are in present continous tense.}
\item \textbf{Prompt \#5}: \textit{There are five transcriptions hypotheses for a given audio and you need to report the true transcription by using english grammar rules.}
\end{itemize}
These prompts are designed based on our experiences with modern LLMs such as GPT \cite{OpenAI2024}, Claude \cite{Anthropic2023} and LLaMA \cite{Touvron2023}.
In Prompt \#1, we tell LLMs that the task is to report the true transcription based on a set of hypotheses. This could bias the LLMs towards re-scoring tasks.
In both Prompt \#1 and \#2, we explicitly inform LLMs that the task is a hard problem. We empirically observe that this often leads to better performance in many tasks, not just limited to post-ASR error correction.
In Promp \#3, we ask the model to summarize the given hypotheses into one sentence, as summarization is a task closely-related to post-ASR error correction.
In Prompt \#4, we explicitly tell LLMs that the data we are dealing with are related to finance, which is true in the case of CHiME-4 \cite{Barker2015, Vincent2016, Barker2017}, as CHiME-4 is built upon the WSJ dataset \cite{Paul1992}.
In Prompt \#5, we ask LLMs to provide gramatically-correct outputs. This could be helpful for post-ASR error correction, as we observe that some ASR transcriptions contain grammar errors.
 
\subsection{Employing EvoPrompt for Prompt Optimization}

In many application scenarios, empirically-designed prompts are often not good enough.
Prompt optimization, which aims at automatically finding a prompt that can lead to better performance for considered tasks, has been attracting wide research interests in LLM research. %

\begin{figure*}
  \centering  
  \includegraphics[width=0.75\linewidth]{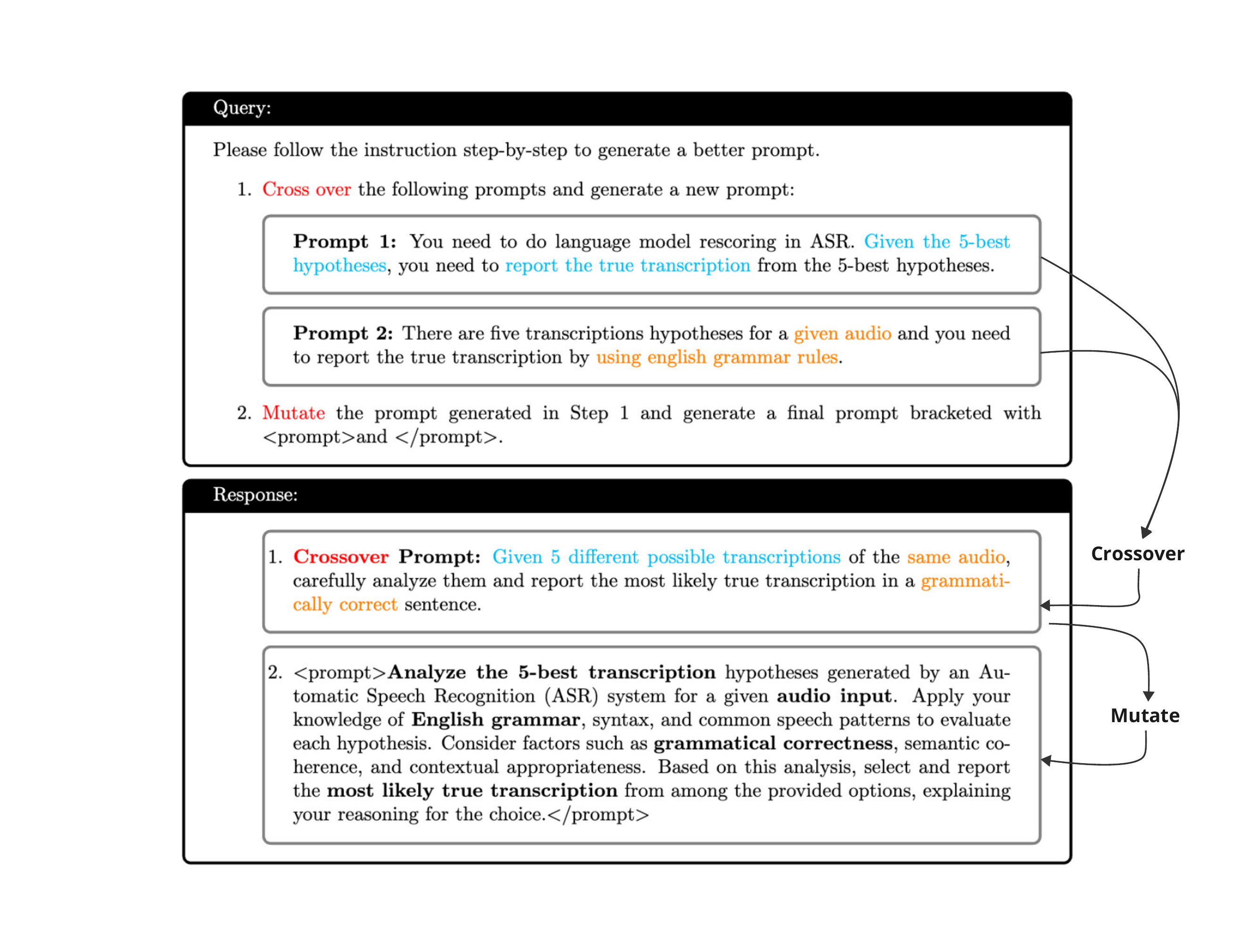}
  \caption{Example of text prompt optimization processes through (i) cross-over and (ii) mutation performed by LLM-operators. %
  }
  \label{cossover_mutation_example}
\end{figure*}

A promising recent algorithm in this direction is EvoPrompt\footnote{\url{https://github.com/beeevita/EvoPrompt}} \cite{Guo2024Evoprompt}, which has shown strong performance and potential on natural language processing tasks such as text classification, simplification and summarization.
EvoPrompt is an iterative prompt optimization algorithm starting from an initial set (or population) of $N$ empirically-designed prompts.
It maintains a set of candidate prompts (initialized with the $N$ empirically-designed prompts) and gradually grows the candidate set by leveraging evolutionary algorithms.
At each evolution iteration, it (a) randomly selects $N$ prompts that have the best scores (denoted as ``best subset'') from the candidate set; (b) creates $N$ new prompts based on the best subset of prompts; (c) checks the score of each new prompt for the targeted tasks; and (d) adds the new prompts and their scores to the maintained candidates set (whose size is now increased by $N$) for the next iteration.
One of the key steps is in how to create each of the new prompts based on the best subset.
In EvoPrompt, each new prompt is created based on two prompts randomly selected from the best subset of prompts, and LLMs are utilized to first \textit{cross-over} the two prompts and then \textit{mutate} the resulting prompt.
See Fig. \ref{cossover_mutation_example} for an example.
This evolutionary mechanism has been shown effective by EvoPrompt \cite{Guo2024Evoprompt}.

In this context, to find prompts that can result in better post-ASR error correction, we employ EvoPrompt for prompt optimization.
We choose the Genetic Algorithm (GA) option in EvoPrompt (see Algorithm $2$ of \cite{Guo2024Evoprompt}), and leverage the set of five empirically-designed prompts described in the previous section (i.e., Prompt-\#\{$1,\dots,5$\}) as the initial candidate set of prompts, and run EvoPrompt for three iterations, where, in each iteration, five new prompts are created (that is, we set $T=3$ and $N=5$ in Algorithm $2$ of \cite{Guo2024Evoprompt}).

\section{Experimental Setup}\label{exp_setup}

We validate the proposed algorithms on the task $1$ of the $2024$ SLT Generative Speech Error Correction (GenSEC) challenge.\footnote{More details of the GenSEC challenge can be found at \url{https://sites.google.com/view/gensec-challenge/home}}
The task is on learning a mapping from N-best list of hypotheses (produced by ASR models) to ground-truth speech transcription so that recognition errors can be corrected. 
The challenge is designed based on the HyPoradise dataset \cite{Chen2023HyPoradise}, which consists of pairs of the N-best list and correct transcription of the utterances derived based on multiple public ASR datasets such as LibriSpeech \cite{Panayotov2015} and CHiME-4 \cite{Barker2015, Vincent2016, Barker2017} by using strong ASR models such as Whisper \cite{Radford2023}, and WavLM \cite{Chen2022} based ASR models.

Different from the official challenge configurations, our experimental setup only uses the CHiME-4 subset of HyPoradise for training.
This is mainly out of cost concerns, as the evolutionary algorithm needs to compute a score for every considered prompt in each iteration.
In CHiME-4 \cite{Barker2015, Vincent2016, Barker2017}, there are $8,738$, $1,640$ and $1,320$ utterances in the training, validation and test set, respectively, and HyPoradise \cite{Chen2023HyPoradise} provides five hypotheses for each utterance.
The task is to predict the true transcription based on the five hypotheses.
In this study, since our main goal is prompt optimization, our system does not leverage acoustic information in error correction and only exploits text-level information.

Besides evaluating optimized prompts on the test set of CHiME-4, we investigate the generalizability of the optimized prompts to unseen domains by using two other datasets (including Common Voice (CV) \cite{Ardila2020} and Wall-Street Jounral (WSJ) \cite{Paul1992}) for evaluation.
There are $836$ test utterances in WSJ and $2,000$ test utterances in CV, and HyPoradise provides five hypotheses for each of the test utterances.

\begin{figure*}
  \centering  
  \includegraphics[width=17cm]{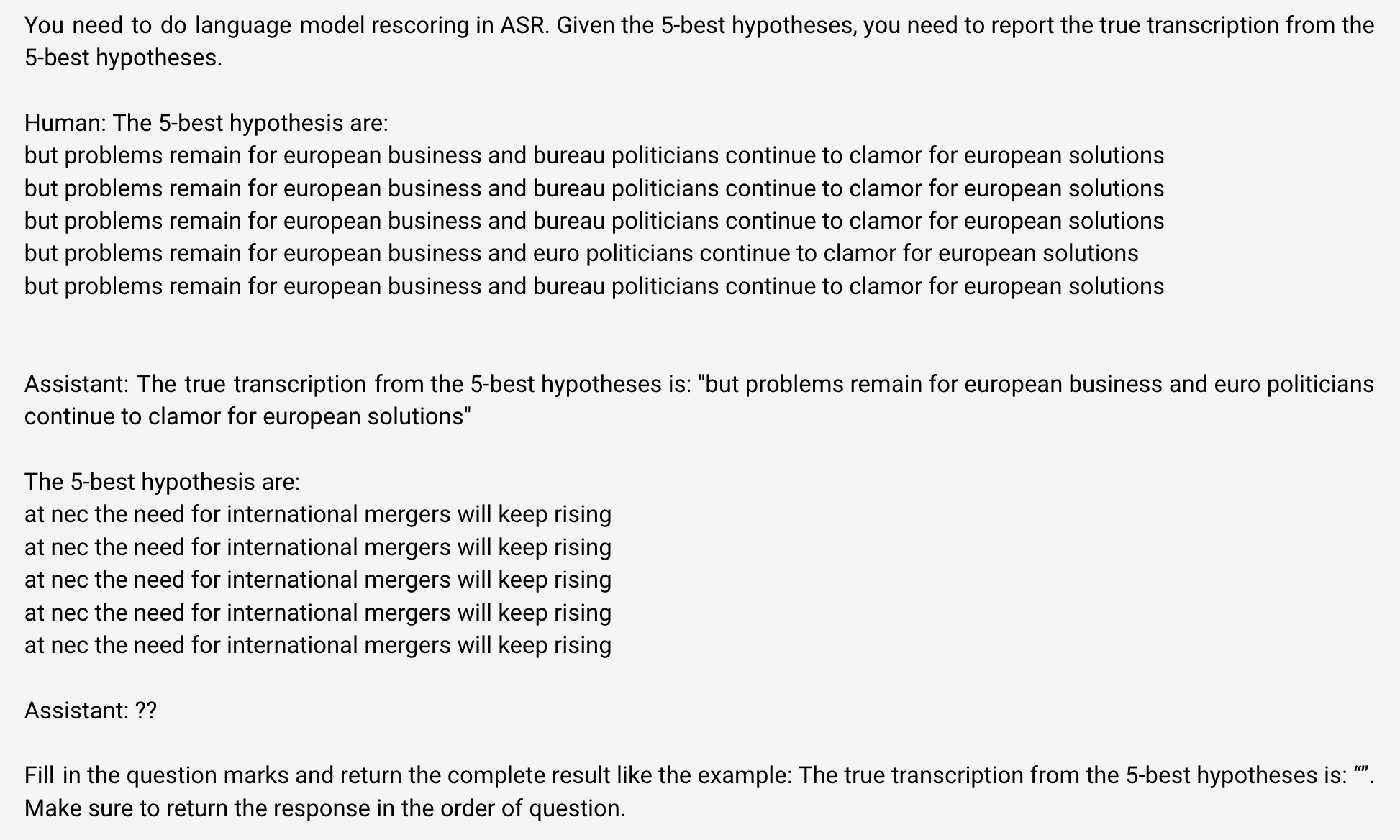}
  \caption{Example input for in-context learning with one demonstration, where the first paragraph denotes the prompt, second and third denotes one demonstration example, and fourth and fifth requests LLMs to correct errors.}
  \label{in_context_learning_example}
\end{figure*}

The LLM we use for prompt optimization is Claude \cite{Anthropic2023}\footnote{We use Claude 3.5 Sonnet. See details in \url{https://www.anthropic.com/news/claude-3-5-sonnet}.}, which has shown competitive performance compared with many state-of-the-art LLMs.
We provide one demonstration example \cite{Min2022} to the LLM before asking the model to perform error correction.
We find that this demonstration strategy is very helpful for in-context learning.
In Fig. \ref{in_context_learning_example}, we provide an example of the input we use in our experiments to LLMs.

We use word error rates (WER) as the evaluation metric, where we optimize our prompt on the training set of the Hyporadise and report WER performance on the unseen test set based on the prompted-LLM  performance on the training set.

\section{Evaluation Results}\label{results}

\subsection{Results of Alternative Prompts}

In row $1$a-$1$e of Table \ref{chime4_results}, we report the results of the proposed alternative prompts on the CHiME-4 dataset.
We observe that the performance is better than the baseline prompt in row $0$, indicating the effectiveness of the proposed alternative prompts.

\begin{table}[]
\scriptsize
\centering
\sisetup{table-format=2.2,round-mode=places,round-precision=2,table-number-alignment = center,detect-weight=true,detect-inline-weight=math}
\caption{Results of post-ASR error correction on CHiME-4 test set}
\label{chime4_results}
\setlength{\tabcolsep}{4pt}
\begin{tabular}{
c %
c %
c %
c %
S[table-format=1.2,round-precision=2] %
}
\toprule
Row & Prompts & \#Demonstrations & \#Iterations & {WER (\%)$\downarrow$} \\
\midrule
$0$ & Baseline Prompt & 0 & - & 7.491788749\\

\midrule

$1$a & Prompt \#$1$ & 0 & - & 7.207486686 \\
$1$b & Prompt \#$2$ & 0 & - & \bfseries 6.390008764 \\
$1$c & Prompt \#$3$ & 0 & - & 6.961429088 \\
$1$d & Prompt \#$4$ & 0 & - & 6.523711287 \\
$1$e & Prompt \#$5$ & 0 & - & 6.508319115 \\

\midrule

$2$a & Prompt \#$1$ & 1 & - &  \bfseries 5.99245126932423 \\
$2$b & Prompt \#$2$ & 1 & - & 6.209606535339434 \\
$2$c & Prompt \#$3$ & 1 & - & 6.359547076159454 \\
$2$d & Prompt \#$4$ & 1 & - & 6.147443255260844 \\
$2$e & Prompt \#$5$ & 1 & - & 6.2251176257690916 \\

\midrule

$3$a & Mutated Prompt \#$1$ & 1 & 1 & \bfseries 6.3078434 \\
$3$b & Mutated Prompt \#$2$ & 1 & 1 & 9.1356994\\
$3$c & Mutated Prompt \#$3$ & 1 & 1 & 8.0554263\\
$3$d & Mutated Prompt \#$4$ & 1 & 1 & 9.2703771\\
$3$e & Mutated Prompt \#$5$ & 1 & 1 & 8.4173517\\

\midrule

$4$a & Mutated Prompt \#$1$ & 1 & 2 & 7.238509 \\
$4$b & Mutated Prompt \#$2$ & 1 & 2 & \bfseries 5.377178 \\
$4$c & Mutated Prompt \#$3$ & 1 & 2 & 6.169429 \\
$4$d & Mutated Prompt \#$4$ & 1 & 2 & 8.242739 \\
$4$e & Mutated Prompt \#$5$ & 1 & 2 & 5.785637 \\

\midrule

$5$a & Mutated Prompt \#$1$ & 1 & 3 & 6.297503 \\
$5$b & Mutated Prompt \#$2$ & 1 & 3 & \bfseries 4.875652 \\
$5$c & Mutated Prompt \#$3$ & 1 & 3 & 6.219947 \\
$5$d & Mutated Prompt \#$4$ & 1 & 3 & 5.340011 \\
$5$e & Mutated Prompt \#$5$ & 1 & 3 & 6.406081 \\

\bottomrule
\end{tabular}
\end{table}

\subsection{Results of EvoPrompt for Prompt Optimization}

In $2$a-$2$e of Table \ref{chime4_results}, we provide one demonstration example before asking LLMs to do error correction.
See Fig. \ref{in_context_learning_example} for an example.
Comparing $2$a-$2$e with $1$a-$1$e, we observe that including one demonstration improves the performance.

In $2$a-$4$e of Table \ref{chime4_results}, we report the results of using EvoPrompt to optimize the set of empirically-designed prompts (\textit{i.e.}, Prompt-\#\{$1,\dots,5$\}).
We provide the WER of each of the five new (i.e., mutated) prompts in each iteration.
We observe that the performance gets gradually better with more iterations.
The prompt in row $5$b stands out with the lowest WER of $4.88$\%.
This indicates the effectiveness and potential of EvoPrompt for prompt optimization.

\subsection{Analysis of Optimized Prompt}

The optimized prompt obtained by applying EvoPrompt to Prompt-$1$ in the first iteration is: \textit{You are presented with 5 different transcription hypotheses for a single audio clip from a financial meeting. Your task is to critically evaluate these options, considering factors such as context, coherence, and English language conventions. Synthesize the most likely and accurate representation of the audio content into one concise, grammatically correct sentence that is mostly present continuous tense}.
Comparing it with Prompt \#1 presented in Section \ref{alterative_prompts}, we observe that:
\begin{itemize}[leftmargin=*,noitemsep,topsep=0pt]
\item The mutated prompt provides clearer instructions, which facilitate the task for the LLM. Instead of a vague instruction like \textit{report the true transcription}, an optimized instruction provides specific guidelines, such as \textit{Your task is to critically evaluate these options, taking into account factors like context, coherence, and English language conventions}.
\item The improved prompts are more appropriate to the context, ensuring that they are tailored to the dataset. For example, for the CHiME-4 dataset, which contains financial transcriptions, the mutated prompt is focused on financial terms and contexts.
\item Including one demonstration example in the layout of the prompts significantly helps the LLM understand the desired output style and how to correct errors. Demonstrations can provide a clear reference, reducing ambiguity, and guiding the LLM towards producing more accurate transcriptions.
\end{itemize}

\subsection{Examples of Corrected Errors}

Table \ref{error_correction} provides an example of the corrected errors.
We observe that biasing LLMs towards the domain of finance helps correct errors such as ``fidelity'' and ``gencorp'', which are financial terms.

\begin{table}[h]
\scriptsize
\centering
\caption{Example of corrected errors based on an utterance drawn from CHiME-4. Best viewed in color.}
\label{error_correction}
\resizebox{1.0\columnwidth}{!}{
\begin{tabular}{@{}cp{4cm}c@{}}

\toprule

Type & Utterance & WER (\%)$\downarrow$ \\

\midrule

\multirow{2}{*}{1\textsuperscript{st} Hypo. by AM} & \textcolor{red}{fatelli} had contended that \textcolor{red}{genecorp} is not a qualified broadcaster because it failed to disclose allegedly improper political campaign contributions and foreign payments & \multirow{2}{*}{$8.69$} \\

\midrule

\multirow{2}{*}{2\textsuperscript{nd} Hypo. by AM} & fidelity had contended that \textcolor{red}{jeane corp} is not a qualified broadcaster because it failed to disclose allegedly improper political campaign contributions and foreign payments & \multirow{2}{*}{$8.69$} \\

\midrule

\multirow{2}{*}{Correction by LLM} & \textcolor{blue}{fidelity} had contended that \textcolor{blue}{gencorp} is not a qualified broadcaster because it failed to disclose allegedly improper political campaign contributions and foreign payments & \multirow{2}{*}{$0$} \\

\midrule

\multirow{2}{*}{Ground-truth Transcription} & \textbf{fidelity} had contended that \textbf{gencorp} is not a qualified broadcaster because it failed to disclose allegedly improper political campaign contributions and foreign payments & \multirow{2}{*}{-} \\

\bottomrule

\end{tabular}
}
\end{table}

\subsection{Generalizability of Optimized Prompts to Unseen Domains}

Table \ref{cv_test_results} and \ref{wsj_test_results} report the results on the CV and WSJ test sets.
They offer valuable understandings about the ability of the model to extend its performance beyond the data it is trained on.
The best performing mutated prompt \#$1$ from the CHiME experiement shows a deterioration in performance on both test sets, with a higher WER compared to the baseline. This implies that the mutations implemented in Prompt \#1 are not beneficial for the process of generalization. 

The mutated prompt \#$2$ from the second and third iteration of the experiment shows improved performance ($2.79$\% vs. $2.84$\%) on the WSJ test set and a slight improvement ($10.92$\% vs. $10.71$\%) on the CV test set. This indicates that the specific alterations in prompt may have contributed generalized traits that benefit both seen and unseen areas.

Therefore, the findings indicate that the ability of post-ASR error correcting models to make generalizations might vary greatly depending on the initial prompts employed (i.e., Prompt-\#\{$1,\dots,5$\}). Further research could explore the specific words and characteristics of the prompt that contribute to its generalizability.

\begin{table}[]
\scriptsize
\centering
\sisetup{table-format=2.2,round-mode=places,round-precision=2,table-number-alignment = center,detect-weight=true,detect-inline-weight=math}
\caption{Results of post-ASR error correction using evolutionary prompts on unseen Common Voice test set of Hyporadise
}
\label{cv_test_results}
\setlength{\tabcolsep}{4pt}
\begin{tabular}{
c %
c %
c %
S[table-format=1.2,round-precision=2] %
}
\toprule
Row & Prompts & \#Demonstrations & {WER (\%)$\downarrow$} \\

\midrule

0 & Baseline Prompt & 0 & 11.0628604 \\
\midrule

$3$a & Mutated Prompt \#$1$ & 0 & 13.56161458 \\
\midrule

$4$b & Mutated Prompt \#$2$ & 0  & 10.91913327 \\
\midrule

$5$b & Mutated Prompt \#$2$ & 0 & \bfseries 10.71 \\
\bottomrule
\end{tabular}
\end{table}

\begin{table}[]
\scriptsize
\centering
\sisetup{table-format=2.2,round-mode=places,round-precision=2,table-number-alignment = center,detect-weight=true,detect-inline-weight=math}
\caption{Results of post-ASR error correction using evolutionary prompts on unseen Wall-Street Journal test set of Hyporadise}
\label{wsj_test_results}
\setlength{\tabcolsep}{4pt}
\begin{tabular}{
c %
c %
c %
S[table-format=1.2,round-precision=2] %
}
\toprule
Row & Prompts & \#Demonstrations & {WER (\%)$\downarrow$} \\

\midrule

0 & Baseline Prompt & 0 & 3.67 \\
\midrule

$3$a & Mutated Prompt \#$1$ & 0 & 14.74883178 \\
\midrule

$4$b & Mutated Prompt \#$2$ & 0  & \bfseries 2.79 \\
\midrule
$5$b & Mutated Prompt \#$2$ & 0 & 2.840245327 \\

\bottomrule
\end{tabular}
\end{table}

\subsection{Cost of Proposed Algorithms}

The API for Claude 3.5 Sonnet is charged at \$3 per million input tokens and \$15 per million output tokens.
It provides more input and output tokens at a reduced cost than the previous Claude 3 Opus model.
The cost for a single experiment on the CHiME-4 dataset using Claude Sonnet 3.5 is around \$$2.2$ and for the complete experiment the cost is around \$$60.0$.
Check out the pricing page of Claude\footnote{\url{https://www.anthropic.com/pricing\#anthropic-api}} for more details.

\section{Conclusions}\label{conclusions}

We have proposed alternative prompts for post-ASR error corrections, and leveraged an evolutionary prompt optimization algorithm for prompt optimization.
Evaluation results on the CHiME-4 subset of the GenSEC Challenge Task $1$ show the effectiveness of the proposed algorithms, suggesting that leveraging conventional evolutionary algorithms for prompt optimization is a promising research direction.
Moving forward, we plan to investigate LLM fine-tuning to find better prompts for generative error correction.
\subsubsection*{Acknowledgment}
The authors would like to thank Prof. Shinji Wanatabe for his valuable feedbacks and encouragements on experimental designs when initializing the project. 
\clearpage
{\small
\bibliographystyle{IEEEtran}
\bibliography{references.bib}
}

\end{document}